\documentclass[letterpaper]{article} 
\usepackage[]{aaai2026}  
\usepackage{times}  
\usepackage{helvet}  
\usepackage{courier}  
\usepackage[hyphens]{url}  
\usepackage{graphicx} 
\usepackage{natbib}  
\usepackage{caption} 
\usepackage{xcolor}
\usepackage{threeparttable}
\usepackage{booktabs}
\definecolor{ForestGreen}{RGB}{34,139,34}
\definecolor{BrickRed}{RGB}{178,34,34}
\usepackage{multirow}
\usepackage{amsmath} 
\usepackage{booktabs}
\usepackage{algorithm}
\usepackage{algorithmic}

\usepackage{newfloat}
\usepackage{listings}
\DeclareCaptionStyle{ruled}{labelfont=normalfont,labelsep=colon,strut=off} 
\floatstyle{ruled}
\newfloat{listing}{tb}{lst}{}
\floatname{listing}{Listing}
\title{Error Correction in Radiology Reports: A Knowledge Distillation-Based Multi-Stage Framework}

\author{
    Jinge Wu\textsuperscript{\rm 1}\equalcontrib,
    Zhaolong Wu\textsuperscript{\rm 2}\equalcontrib,
    Ruizhe Li\textsuperscript{\rm 3},
    Tong Chen\textsuperscript{\rm 4},
    Abul Hasan\textsuperscript{\rm 5},
    Yunsoo Kim\textsuperscript{\rm 1},\\
    Jason Pui-Yin Cheung\textsuperscript{\rm 2}\correspondingauthor,
    Teng Zhang\textsuperscript{\rm 2}\correspondingauthor,
    Honghan Wu\textsuperscript{\rm 1,6}\correspondingauthor
}

\affiliations{
\textsuperscript{\rm 1}University College London,
\textsuperscript{\rm 2}The University of Hong Kong,
\textsuperscript{\rm 3}University of Aberdeen,\\
\textsuperscript{\rm 4}The University of Sydney,
\textsuperscript{\rm 5}University of Oxford,
\textsuperscript{\rm 6}University of Glasgow\\
honghan.wu@ucl.ac.uk,
\{cheungjp, tgzhang\}@hku.hk \\
}

\begin{document}
\maketitle

\begin{abstract}
The increasing complexity and workload of clinical radiology leads to inevitable oversights and mistakes in their use as diagnostic tools, causing delayed treatments and sometimes life-threatening harm to patients. While large language models (LLMs) have shown remarkable progress in many tasks, their utilities in detecting and correcting errors in radiology reporting are limited. This paper proposes a novel dual-knowledge infusion framework that enhances LLMs' capability for radiology report proofreading through systematic integration of medical expertise.
Specifically, the knowledge infusion combines medical knowledge graph distillation (MKGD) with external knowledge retrieval (EXKR), enabling an effective automated approach in tackling mistakes in radiology reporting. By decomposing the complex proofreading task into three specialized stages of detection, localization, and correction, our method mirrors the systematic review process employed by expert radiologists, ensuring both precision and clinical interpretability. To perform a robust, clinically relevant evaluation, a comprehensive benchmark is also proposed using real-world radiology reports with real-world error patterns, including speech recognition confusions, terminology ambiguities, and template-related inconsistencies. Extensive evaluations across multiple LLM architectures demonstrate substantial improvements of our approach: up to 31.56\% increase in error detection accuracy and 37.4\% reduction in processing time. Human evaluation by radiologists confirms superior clinical relevance and factual consistency compared to existing approaches. 
\end{abstract}

\begin{links}

\link{code}{https://github.com/knowlab/MedKIC-Radiology-Proofreading}

\end{links}


\section{Introduction}

\label{sec:intro}
\begin{figure*}[ht]
  \centering
  \includegraphics[width=0.9\textwidth,trim={0 100 50 150},clip]{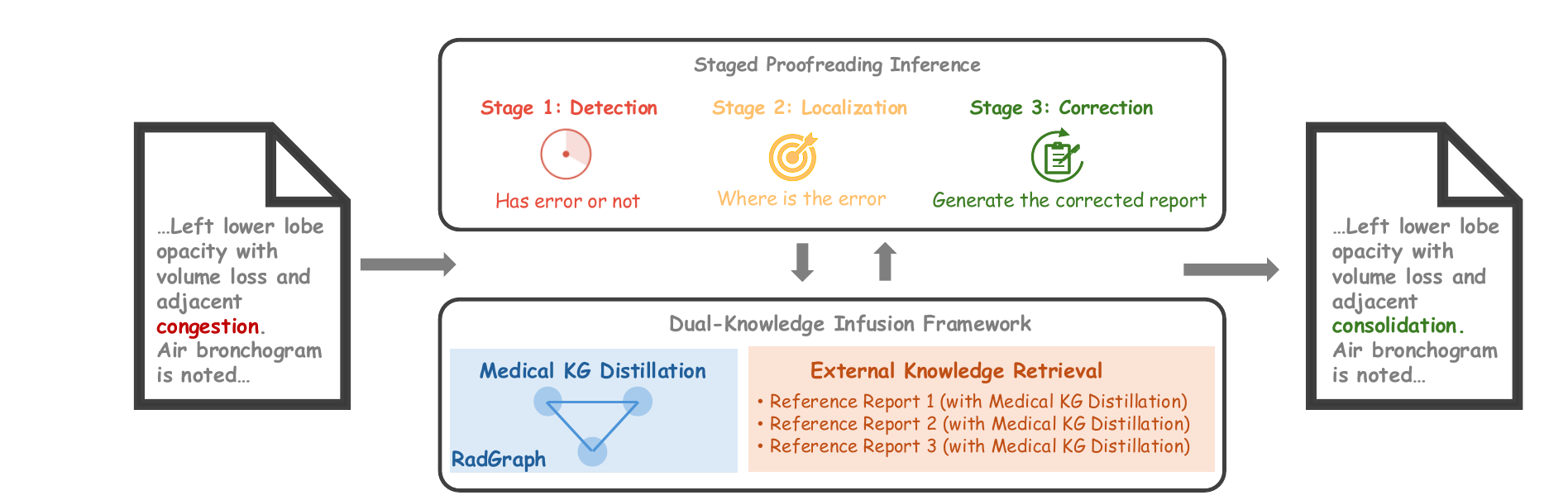}
  \caption{An overview of our medical report proofreading framework. The staged inference process (top) breaks down error correction into detection (identifying error presence), localization (pinpointing error terms like ``congestion"), and correction (providing proper replacements like ``consolidation"). The dual-knowledge infusion framework (bottom) supports this process through MKGD's structural analysis and EXKR's domain knowledge integration, enabling accurate and clinically sound corrections.}
  \label{fig:label0}
\end{figure*} 
The landscape of medical documentation is undergoing rapid transformation, with radiology reports becoming increasingly complex to support precision medicine and comprehensive patient care \cite{brady2017error, wu2022survey, collins2015new, topol2019high}. However, this evolution has introduced significant challenges that threaten documentation quality and reliability, with critical implications for patient safety and healthcare efficiency.

Retrospective studies indicate that approximately 30\% of radiological examinations contain documentation errors, ranging from terminology inconsistencies and negation mistakes to contextual contradictions \cite{kasalak2023work, berlin2007malpractice, wu2024hallucination, wu2023exploring}. These errors manifest through speech recognition misinterpretations (e.g., ``effusion'' becoming ``infusion'') and template-based inconsistencies, leading to delayed diagnoses, inappropriate treatments, and compromised patient safety \cite{pinto2018speech}. The clinical consequences of such errors extend beyond immediate patient care, affecting treatment planning, follow-up decisions, and medicolegal documentation.

Manual review processes are becoming increasingly unsustainable as examination volumes increase by 3-5\% annually while specialist availability grows at less than 2\% per year \cite{sunshine2006radiology, kruskal2011changes, wu2024radbartsum}. This growing workload burden particularly affects busy clinical environments and resource-constrained healthcare facilities, where comprehensive manual proofreading becomes impractical, leading to potential quality compromises in medical documentation.

Current automated approaches face four critical limitations that hinder their clinical adoption and effectiveness. Existing methods lack systematic medical knowledge integration, operating with linguistic correctness that ignores clinical appropriateness---unable to distinguish between medically distinct terms like ``consolidation'' and ``atelectasis'' \cite{huang2019clinicalbert, wang2018clinical, wu2024hybrid}. They provide black-box decision-making incompatible with clinical environments where practitioners must validate recommendations \cite{cabitza2017unintended, sendak2020human}. Additionally, these approaches treat error correction as monolithic tasks without modeling expert cognitive processes, and rely on static training paradigms that cannot adapt to evolving medical knowledge like COVID-19 terminology \cite{ng2020imaging}---limiting their utility in dynamic clinical practice.

To address these challenges, we present a novel framework combining staged proofreading inference with dual-knowledge infusion, as illustrated in Figure 1. Our staged approach decomposes error correction into detection, localization, and correction phases, mirroring expert review processes while enabling transparent decision-making that can be validated by clinical practitioners. Our dual-knowledge mechanism combines medical knowledge graph distillation (MKGD) with external knowledge retrieval (EXKR), transforming clinical reports into structured representations and leveraging clinically relevant reference cases without expensive fine-tuning---providing scalable quality assurance that enhances rather than replaces clinical expertise.

Our work makes four key contributions with demonstrated benefits for clinical practice: 1) \textbf{Methodological Innovation}---staged inference framework enabling transparent AI-assisted decision support that integrates seamlessly with clinical workflows, 2) \textbf{Knowledge Integration Architecture}---dual-knowledge infusion for scalable domain adaptation that leverages existing medical knowledge without requiring extensive retraining, 3) \textbf{Clinical Validation Framework}---comprehensive benchmark with authentic error patterns validated by practicing radiologists, ensuring real-world applicability and clinical relevance, and 4) \textbf{Demonstrated Clinical Impact}---significant improvements (up to 31.56\% in error detection, 37.4\% processing time reduction) that directly enhance report quality and reduce radiologist workload, particularly benefiting high-volume clinical environments where efficiency gains translate to improved patient care.


\section{Related Work}

Research in automated medical text error detection reveals critical limitations that highlight the need for more sophisticated clinical solutions.

\textbf{General Text Correction Approaches} achieve success in linguistic domains \cite{fang2024llmcl, kamoi2024evaluating} but fail in medical contexts where clinical appropriateness supersedes linguistic correctness. While correcting ``recieve" to ``receive," they cannot distinguish between ``consolidation" and ``atelectasis"—legitimate medical terms representing different pathophysiological processes requiring distinct treatments.

\textbf{Medical Domain-Specific Approaches} like MEDIQA-CORR 2024 \cite{ben2024overview} focus on general clinical notes using synthetic data rather than specialized radiology reports. Recent medical language models show promise through fine-tuning \cite{zhou2023survey, abacha2023empirical} but operate as black boxes incompatible with clinical environments where radiologists must validate automated recommendations.

\textbf{Knowledge-Enhanced Methods} incorporate structured medical knowledge like UMLS concepts \cite{rajpurkar2022ai} but treat error correction as monolithic tasks without modeling expert cognitive processes. A radiologist would systematically: (1) assess error presence, (2) identify ``congestion" as inappropriate in pulmonary context, and (3) correct to ``consolidation." Current methods cannot decompose this reasoning chain.

\textbf{Static Training Limitations} present the most fundamental challenge. Current approaches freeze medical knowledge at training time, creating vulnerabilities in rapidly evolving practice \cite{lazaridou2021mind, sun2025generative, singhal2023large}. Models struggle with novel terminology like ``COVID pneumonia" and cannot incorporate evolving guidelines without expensive retraining.

\begin{figure*}[ht]
  \centering
  \includegraphics[width=0.9\textwidth,trim={20 60 20 50},clip]{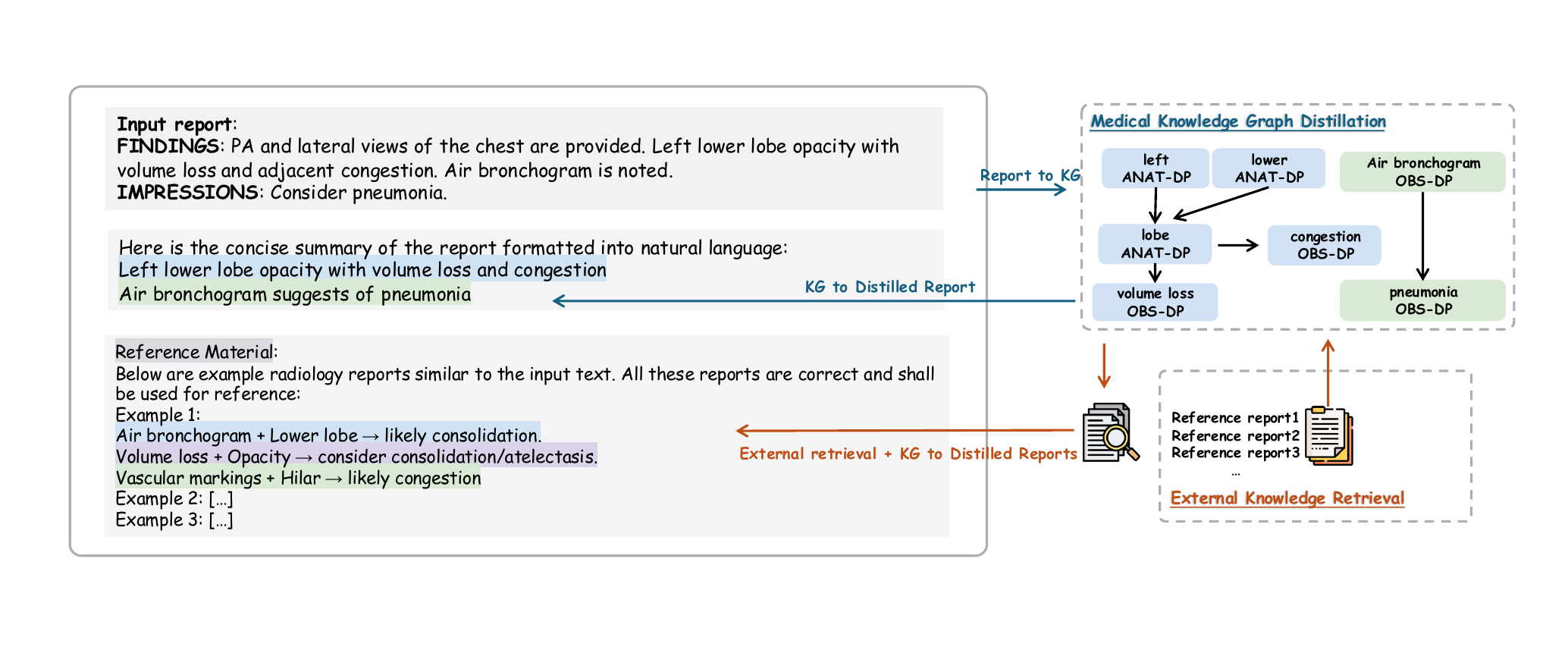}
  \caption{Illustration of our dual-knowledge infusion framework. Left: Input medical report with task description and reference examples. Right: MKGD transforms the report into a structured graph representation capturing anatomical entities (ANAT-DP) and observations (OBS-DP/DA) with their relationships (modify, located\_at, suggestive\_of), while EXKR provides relevant domain knowledge from reference reports to guide the correction process.}
  \label{fig:label2}
\end{figure*} 

\section{Methodology}

\subsection{Staged Proofreading Inference}
We decompose error correction task into three specialized stages that mirror expert radiologist review processes, enabling transparent and clinically-grounded reasoning.

\begin{itemize}
\item \textbf{Stage 1: Error Detection} performs binary classification to identify whether a report contains errors. The model analyzes global patterns and medical consistency indicators to distinguish between correct and problematic reports, focusing on medical-specific characteristics that indicate potential documentation issues

\item \textbf{Stage 2: Error Localization} conducts fine-grained analysis to pinpoint specific error locations within reports identified as problematic. This stage examines individual medical entities, their relationships, and broader clinical context to identify the precise terms or phrases requiring correction.

\item \textbf{Stage 3: Error Correction} generates clinically appropriate corrections based on detected error types and locations. The correction process is guided by both local context and broader medical knowledge to ensure medical accuracy and report consistency while maintaining clinical validity.
\end{itemize}

Each stage is designed with medical-specific reasoning mechanisms that enable more precise and reliable error handling, while maintaining the broader clinical context. This structured decomposition allows the model to focus on specific aspects of medical error analysis at each stage, leading to more accurate and interpretable results compared to generic language model applications. Moreover, this step-by-step approach helps reduce hallucination and encourages more targeted corrections that align with medical knowledge and practices.

\subsection{Dual-Knowledge Infusion Framework}

Detecting domain-specific errors requires bridging local textual coherence with medical knowledge that extends beyond the immediate document. To that end, we propose a \emph{dual-retrieval} mechanism comprised of medical knowledge graph distillation (MKGD) with external knowledge retrieval (EXKR), as shown in Figure 2.

\subsubsection{Medical Knowledge Graph Distillation (MKGD)}

We transform clinical reports into structured medical entity graphs using RadGraph \cite{jain2021radgraph}, a specialized framework designed for clinical information extraction from radiology reports. RadGraph has been extensively validated and demonstrates high performance in extracting clinically relevant entities and relationships from chest X-ray reports.

\textbf{Entity Extraction} identifies two main categories that capture the fundamental components of radiological descriptions. Anatomical entities (ANAT) represent specific body structures mentioned in reports, encompassing organs, tissues, and anatomical landmarks such as lungs, ribs, cardiac structures, and vascular components. Observational entities (OBS) capture clinical findings and are systematically classified into three distinct certainty levels that preserve diagnostic confidence: Definitely Present (DP) indicates confirmed findings with high clinical certainty, Uncertain (U) represents possible or equivocal findings requiring further evaluation, and Definitely Absent (DA) explicitly documents ruled-out conditions or normal findings. This classification system ensures that the extracted knowledge graph preserves the nuanced diagnostic reasoning inherent in radiological interpretations.

\textbf{Relationship Modeling} captures three fundamental relation types that form the semantic structure of medical knowledge. The \textit{suggestive of} relation encodes diagnostic reasoning chains by linking observational findings to their potential clinical implications, such as when specific imaging patterns suggest particular pathological conditions. The \textit{located at} relation establishes precise anatomical localization by connecting observational findings to their corresponding anatomical sites, enabling accurate spatial mapping of clinical findings within body structures. The \textit{modify} relation captures hierarchical and descriptive relationships between entities, including anatomical modifiers that specify spatial characteristics (e.g., ``left", ``lower", ``bilateral") and observational qualifiers that describe finding attributes such as severity (``mild", ``severe") or morphological characteristics (``nodular", ``linear").

\textbf{Graph-to-Text Conversion} transforms the structured representation back into human-readable sentences through systematic rule-based integration that preserves both clinical accuracy and linguistic naturalness. The conversion process operates through several coordinated phases: entity classification categorizes nodes by type and certainty level, semantic integration combines related entities into meaningful phrases through hierarchical sorting of modifiers according to clinical conventions (e.g., combining ⟨lower, modify, lobe⟩ and ⟨opacity, located at, lobe⟩ into ``lower lobe opacity"), logical reasoning applies domain-specific transformations with attention to negations and uncertainty indicators (converting ⟨opacity, located at, lobe⟩ with DA certainty to ``no lobe opacity"), and sentence construction groups phrases by anatomical regions while applying grammatical rules to generate well-formed sentences. This systematic approach ensures that the generated natural language maintains both clinical accuracy and readability while preserving the semantic richness of the original knowledge graph representation.

\subsubsection{External Knowledge Retrieval (EXKR)} 
In addition to knowledge graph distillation, our method integrates knowledge from external clinical sources by leveraging a curated database of error-free reference reports to provide real-world expertise and domain-specific knowledge patterns. This external knowledge component addresses the limitation of analyzing reports in isolation by incorporating broader clinical experience and established medical knowledge patterns.

\textbf{Reference Selection} employs semantic similarity matching using e5-large-unsupervised embeddings to identify the top-k most relevant reference reports based on cosine similarity scores with the input report. We empirically determined that k=4 provides optimal performance across different model architectures while maintaining computational efficiency. This similarity-based approach ensures that retrieved references are topically relevant and contextually appropriate for the specific clinical scenario under analysis.

\textbf{Knowledge Standardization} processes selected reference reports through the same entity extraction and graph-to-text conversion pipeline used for input reports, creating structured natural language statements that provide domain-specific knowledge rules and clinical patterns. This standardized processing ensures consistent knowledge representation while enabling direct comparison between input reports and reference examples. The resulting knowledge statements capture both explicit medical relationships (e.g., ``air bronchogram + lower lobe → likely consolidation") and implicit clinical patterns derived from validated practice.

\begin{table*}[t]
\centering
\scriptsize
\resizebox{\textwidth}{!}{%
\begin{tabular}{lllllllll}
\toprule
\multirow{3}{*}{\textbf{Model}} & \multicolumn{2}{c}{\textbf{Error Detection (Acc\%)}} & \multicolumn{2}{c}{\textbf{Error Localization (Acc\%)}} & \multicolumn{3}{c}{\textbf{Error Correction (AggNLG)}} \\
\cmidrule(lr){2-3} \cmidrule(lr){4-5} \cmidrule(lr){6-8}
& \textbf{Baseline} & \textbf{Our Method} & \textbf{Baseline} & \textbf{Our Method}&  \textbf{Baseline} &  \textbf{Baseline} & \textbf{Our Method} \\
&  & \textbf{(MKGD+EXKR)} &  & \textbf{(MKGD+EXKR)} &\textbf{(End-to-End)} &\textbf{(Staged)} & \textbf{(MKGD+EXKR)} \\
\midrule
\multicolumn{8}{l}{\textit{Medical Domain Models}} \\
MMedLM2 & 41.49 & \textbf{73.05} \textcolor{ForestGreen}{(+31.56)} & 30.94 & \textbf{46.05} \textcolor{ForestGreen}{(+15.11)} & 47.80 & 58.27 & \textbf{53.50} \textcolor{ForestGreen}{(-5.70)} \\
Llama3-Aloe & 45.31 & \textbf{67.26} \textcolor{ForestGreen}{(+21.95)} & 43.34 & \textbf{51.35} \textcolor{ForestGreen}{(+8.01)} & 63.33 & 90.17 & \textbf{74.77} \textcolor{ForestGreen}{(+11.44)} \\
\midrule
\multicolumn{8}{l}{\textit{General Purpose Models}} \\
Phi3-mini & 67.26 & \textbf{73.06} \textcolor{ForestGreen}{(+5.80)} & 47.71 & \textbf{52.65} \textcolor{ForestGreen}{(+4.94)}  & 74.36 & 74.08 & \textbf{78.85} \textcolor{ForestGreen}{(+4.49)} \\
Phi3-small & 79.03 & \textbf{80.21} \textcolor{ForestGreen}{(+1.18)} & 63.44 & \textbf{65.04} \textcolor{ForestGreen}{(+1.60)}  & 80.03 & 86.57 & \textbf{86.67} \textcolor{ForestGreen}{(+6.64)} \\
Phi3-medium & 73.67 & \textbf{79.04} \textcolor{ForestGreen}{(+5.37)} & 69.73 & 63.44 \textcolor{BrickRed}{(-6.29)} & 84.47  & 90.25 & \textbf{92.25} \textcolor{ForestGreen}{(+7.78)} \\
Llama3-8B & 37.79 & \textbf{62.27} \textcolor{ForestGreen}{(+24.48)} & 37.29 & \textbf{53.14} \textcolor{ForestGreen}{(+15.85)} & 84.34 & 94.29 & \textbf{94.43} \textcolor{ForestGreen}{(+10.09)} \\
\midrule
\textbf{Average} & {57.43} & \textbf{72.48} \textcolor{ForestGreen}{\textbf{(+15.05)}} & {48.74} & \textbf{55.28} \textcolor{ForestGreen}{\textbf{(+6.54)}} & {72.39} & {82.27} & \textbf{80.08} \textcolor{ForestGreen}{\textbf{(7.69)}} \\
\bottomrule
\end{tabular}%
}
\caption{Overall Performance Comparison: Results of the staged inference framework with dual-knowledge infusion (MKGD + EXKR). For Error Detection and Error Localization, values represent accuracy (\%). For Error Correction, results are reported using the aggregate NLG score (AggNLG), computed as the average of ROUGE-1, BERTScore, and BLEU. Two baselines are compared: (1) End-to-End, where the model directly generates corrected reports, and (2) Staged Baseline, which applies staged inference without knowledge infusion. Our method adds dual-knowledge infusion (MKGD + EXKR) on top of the staged framework.}
\label{tab:main_comparison}
\end{table*}
\textbf{Contextual Integration} combines the standardized reference knowledge with input report analysis to provide comprehensive medical context. This integration process considers both the structural similarities captured through MKGD representations and the semantic relationships identified through EXKR patterns, enabling detection of subtle inconsistencies that might be missed by either approach alone. The dual-level framework leverages both the internal logic of individual reports and the accumulated wisdom of clinical experience encoded in reference examples.

\subsection{Integration with LLM}

We carefully design prompt engineering strategies that integrate both knowledge sources into natural language instructions, enabling effective communication of complex medical context to language models while maintaining clarity and focus. The prompt template architecture follows a consistent structure across all three stages: (1) professional role definition establishing the LLM as a radiologist with specific expertise in chest radiology and diagnostic report writing, (2) task-specific instructions tailored to the current stage's objectives, clearly specifying the expected analysis focus and decision criteria, (3) structured knowledge integration that presents both MKGD-generated summaries and EXKR reference examples in organized, digestible formats, and (4) explicit output format specifications that ensure consistent responses suitable for downstream processing and clinical validation.

\subsection{Benchmark Data Construction}
\label{sec:data}

As there was no prior study for this task when we started this work, we constructed a comprehensive benchmark that reflected real-world clinical scenarios by introducing real-world scenario derived error patterns into validated radiology reports. This approach ensured that our evaluation captured clinically relevant documentation challenges, providing meaningful assessment of system performance in clinical contexts.

The dataset foundation was the MIMIC-CXR dataset \cite{johnson2019mimic}, a large-scale collection of real-world radiology reports. We constructed two distinct sets: (1) a reference set of 112,251 error-free radiology reports that serve as the knowledge base for our framework's EXKR component, and (2) an evaluation set of 1,622 radiology reports comprising 512 error-free examples for baseline comparison and 1,110 reports containing systematically introduced errors for comprehensive testing. The following three strategies were adopted to introduce errors. Although the original reports are assumed to be error-free, minor real-world inaccuracies may still exist.

\textbf{General Error Introduction Strategy} focused on clinically realistic scenarios that reflect real-world documentation challenges encountered in clinical practice. We introduced one error per report in either Findings or Impression section. This was to ensure that errors could be detected using cross-sectional information available within the report. This constraint ensured that our evaluation focused on text-based error detection capabilities while maintaining clinical plausibility. The single-error constraint also enabled precise attribution of detection and correction performance to specific types and locations of errors.

\textbf{Negation Error Implementation} was implemented by converting positive findings to negative assertions or vice versa, creating errors that could fundamentally alter clinical interpretations. We utilized RadGraph's negation detection capabilities to identify sentences containing explicit negation markers such as ``no," ``without," ``absence of," or ``ruled out." The error generation process  removed these negation indicators to convert negative findings into positive assertions (e.g., ``no pleural effusion" becomes ``pleural effusion"), or conversely added negation markers to positive findings to create false negative statements. This approach ensured that negation errors were introduced in syntactically appropriate locations while preserving overall sentence structure and medical terminology, creating realistic documentation errors that reflect common transcription and template-related mistakes.

\textbf{Entity-Based Clinical Inconsistency Generation} was a strategy to introduce terminology confusions that reflect real-world documentation challenges. Working with practicing radiologists, we identified 12 critical radiological findings that commonly appear in chest X-ray reports: Atelectasis, Cardiomegaly, Consolidation, Edema, Enlarged Cardiomediastinum, Fracture, Lung Lesion, Lung Opacity, Pleural Effusion, Pleural Other, Pneumonia, and Pneumothorax. For each finding, we established categorized replacement alternatives: speech recognition/spelling confusions reflected common transcription errors (e.g., ``effusion" → ``infusion"), terminology ambiguity represented similar medical terms that may be confused in clinical documentation (e.g., ``congestion" → ``consolidation"), template-related terms captured standardized alternatives used in reporting systems (e.g., ``cardiac enlargement" → ``cardiomegaly"), and other clinical conditions represented related but distinct diagnoses that might be inappropriately substituted.

All introduced errors were validated by practicing radiologists to ensure clinical plausibility and realistic documentation challenges, providing a comprehensive evaluation framework that reflects real-world clinical scenarios.

\begin{table*}[t]
\centering
\scriptsize  
\resizebox{\textwidth}{!}{%
\begin{tabular}{lcccccccccccc}
\toprule
\multirow{3}{*}{\textbf{Model}} & \multicolumn{3}{c}{\textbf{Error Detection (Acc\%)}} & \multicolumn{3}{c}{\textbf{Error Localization (Acc\%)}} & \multicolumn{3}{c}{\textbf{Error Correction (AggNLG)}} & \multicolumn{3}{c}{\textbf{Processing Time (s)}} \\
\cmidrule(lr){2-4} \cmidrule(lr){5-7} \cmidrule(lr){8-10} \cmidrule(lr){11-13}
& \textbf{RAG} & \textbf{Ours} & \textbf{$\Delta$} & \textbf{RAG} & \textbf{Ours} & \textbf{$\Delta$} & \textbf{RAG} & \textbf{Ours} & \textbf{$\Delta$} & \textbf{RAG} & \textbf{Ours} & \textbf{Gain\%} \\
\midrule
\multicolumn{13}{l}{\textit{Medical Domain Models}} \\
MMedLM2 & 52.65 & \textbf{73.05} & \textcolor{ForestGreen}{\textbf{+20.40}} & 35.88 & \textbf{46.05} & \textcolor{ForestGreen}{\textbf{+10.17}} & 48.13 & \textbf{53.50} & \textcolor{ForestGreen}{\textbf{+5.37}} & 30.51 & \textbf{22.00} & \textcolor{ForestGreen}{\textbf{27.90}} \\
Llama3-Aloe & 58.81 & \textbf{67.26} & \textcolor{ForestGreen}{\textbf{+8.45}} & 44.51 & \textbf{51.35} & \textcolor{ForestGreen}{\textbf{+6.84}} & 74.28 & \textbf{74.77} & \textcolor{ForestGreen}{\textbf{+0.49}} & 25.60 & \textbf{16.10} & \textcolor{ForestGreen}{\textbf{37.30}} \\
\midrule
\multicolumn{13}{l}{\textit{General Purpose Models}} \\
Phi3-mini & 65.59 & \textbf{73.06} & \textcolor{ForestGreen}{\textbf{+7.47}} & 42.97 & \textbf{52.65} & \textcolor{ForestGreen}{\textbf{+9.68}} & 75.76 & \textbf{78.85} & \textcolor{ForestGreen}{\textbf{+3.09}} & 24.10 & \textbf{15.50} & \textcolor{ForestGreen}{\textbf{35.60}} \\
Phi3-small & 72.51 & \textbf{80.21} & \textcolor{ForestGreen}{\textbf{+7.70}} & 58.93 & \textbf{65.04} & \textcolor{ForestGreen}{\textbf{+6.11}} & 84.56 & \textbf{86.67} & \textcolor{ForestGreen}{\textbf{+2.11}} & 25.00 & \textbf{15.90} & \textcolor{ForestGreen}{\textbf{36.40}} \\
Phi3-medium & 73.13 & \textbf{79.04} & \textcolor{ForestGreen}{\textbf{+5.91}} & 60.91 & \textbf{63.44} & \textcolor{ForestGreen}{\textbf{+2.53}} & 88.87 & \textbf{92.25} & \textcolor{ForestGreen}{\textbf{+3.38}} & 31.40 & \textbf{21.90} & \textcolor{ForestGreen}{\textbf{30.10}} \\
Llama3-8B & 51.84 & \textbf{62.27} & \textcolor{ForestGreen}{\textbf{+10.43}} & 46.92 & \textbf{53.14} & \textcolor{ForestGreen}{\textbf{+6.22}} & 88.34 & \textbf{94.43} & \textcolor{ForestGreen}{\textbf{+6.09}} & 23.20 & \textbf{14.50} & \textcolor{ForestGreen}{\textbf{37.40}} \\
\midrule
\textbf{Average} & \textbf{62.42} & \textbf{72.48} & \textcolor{ForestGreen}{\textbf{+10.06}} & \textbf{48.35} & \textbf{55.28} & \textcolor{ForestGreen}{\textbf{+6.93}} & \textbf{76.66} & \textbf{80.08} & \textcolor{ForestGreen}{\textbf{+3.42}} & \textbf{26.64} & \textbf{17.65} & \textcolor{ForestGreen}{\textbf{34.10}} \\
\bottomrule
\end{tabular}%
}
\caption{Performance and Efficiency Comparison between Simple RAG and Our Method. Our Method refers to the dual-knowledge infusion framework (MKGD + EXKR). $\Delta$ represents absolute improvement in accuracy/score, while Gain\% shows processing time reduction.}
\label{tab:rag_comparison}
\end{table*}

\section{Experiments}

For evaluation, we select diverse state-of-the-art LLMs spanning general and medical domains. From the general domain, we use LLaMA3-8B and Phi3 variants (mini-3.8B, small-7B, medium-14B) \cite{abdin2024phi3}, allowing us to study the impact of model scale. We also evaluate two medical-specific models, Llama3-Aloe-8B-Alpha and MMedLM2 \cite{qiu2024towards,gururajan2024aloe}. These models built on LLaMA3-8B, and have been further fine-tuned on a wide range of medical instructional datasets, synthetic medical data, medical textbooks, medical websites, etc.

Our evaluation follows the three stages of our framework. For the first two stages, we calculate accuracy for error detection and localization. Regarding the error correction stage, which is only triggered when errors are detected, we employ an aggregate Natural Language Generation (NLG) score (AggNLG) that combines ROUGE-1 \cite{lin2004rouge}, BERTScore \cite{zhang2019bertscore}, and BLEU \cite{sellam2020bleurt}:

\begin{equation}
\footnotesize
\text{AggNLG} = \frac{\text{ROUGE-1} + \text{BERTScore} + \text{BLEU} }{3}
\end{equation}
This composite metric has been used in the related work with \citet{abacha2023investigation} and demonstrated strong correlation with human judgment.

Regaring our staged approach, correction is only performed when errors are detected in the first stage. In this case, we compute AggNLG scores only when both the model output and ground truth indicate the presence of errors. The generally high NLG scores observed in our results can be attributed to the nature of radiological report errors, which typically involve modifying only a few key terms while preserving the majority of the report content.

\paragraph{\textbf{Experiment Setups.}} All the experiments are conducted with 4 NVIDIA GeForce RTX 3090 24576MiB. For our LLM experiments, we utilize the AutoModelForCausalLM from the Hugging Face Transformers library. The model is loaded from the specified model\_id, with the torch\_dtype parameter set to torch.bfloat16. 
We set the max\_new\_tokens=300, do\_sample=True, temperature=0.001 and top\_p=0.8, all other hyperparameters remains unchanged as default values.

\paragraph{\textbf{RAG Pipeline.}}
We use the e5-large-unsupervised model \cite{e5} to transform reports into a vectorized database. This process involves applying cosine similarity to find the most similar texts based on the input radiology reports. The experiment uses parameters set to chunk\_size=1000 and chunk\_overlap=100. We choose top k = 4 reports for the RAG evaluation as it in genral achieves the best performance.

\subsection{Overall Performance}

As shown in Table \ref{tab:main_comparison}, our framework consistently outperforms baselines across all three tasks. The largest gains are observed in error detection (+15.05\%) and error localization (+6.54\%), demonstrating that structured, stage-wise reasoning helps models better identify and pinpoint clinically inconsistent expressions.

In the error correction stage, we evaluate three paradigms—an end-to-end baseline, a staged reasoning baseline, and our staged framework with dual-knowledge infusion. Staged inference already improves correction quality over the end-to-end approach by decomposing the task into interpretable reasoning steps. Building upon this, our dual-knowledge design further enhances factual accuracy and contextual coherence, achieving an average improvement of +7.69\% over the staged baseline, with notable gains for Llama3-Aloe (+11.44\%) and Llama3-8B (+10.09\%). Although small declines appear in MMedLM2, qualitative review indicates fewer hallucinations and higher factual precision. Overall, these results confirm that integrating domain knowledge with structured reasoning yields more accurate, interpretable, and clinically trustworthy corrections across diverse model architectures.


\subsection{Comparison with Simple Retrieval}

Table \ref{tab:rag_comparison} compares our dual-knowledge infusion framework against conventional Retrieval-Augmented Generation (RAG) approaches. While simple RAG retrieves entire documents based on surface-level text similarity, our method employs structured knowledge extraction to guide targeted retrieval of clinically relevant information. Our approach consistently outperforms simple RAG across all metrics, achieving average improvements of 10.06\%, 6.93\%, and 3.42\% for detection, localization, and correction tasks respectively. Medical domain models show the largest gains, with MMedLM2 achieving a 20.40\% improvement in error detection. These results validate our hypothesis that structured, relationship-aware knowledge retrieval is more effective than generic document-level retrieval for medical applications, where understanding anatomical relationships and clinical patterns is crucial for accurate error detection.

\subsection{Computational Efficiency Analysis}

In addition, our more sophisticated framework achieves significant efficiency gains, reducing processing time by 27.9\%-37.4\% across all models while maintaining superior performance. This efficiency improvement stems from our structured approach to information processing: MKGD enables focused entity extraction and relationship analysis, while EXKR provides targeted knowledge integration that avoids unnecessary context processing. The average processing time reduction of 8.99 seconds per instance (from 26.64s to 17.65s) demonstrates that structured reasoning can be both more accurate and computationally efficient than brute-force approaches. These efficiency gains are particularly valuable for clinical deployment, where timely analysis is crucial for workflow integration.

\subsection{Human Evaluation}

\begin{figure}[h]
  \centering
  \includegraphics[width=0.48\textwidth]{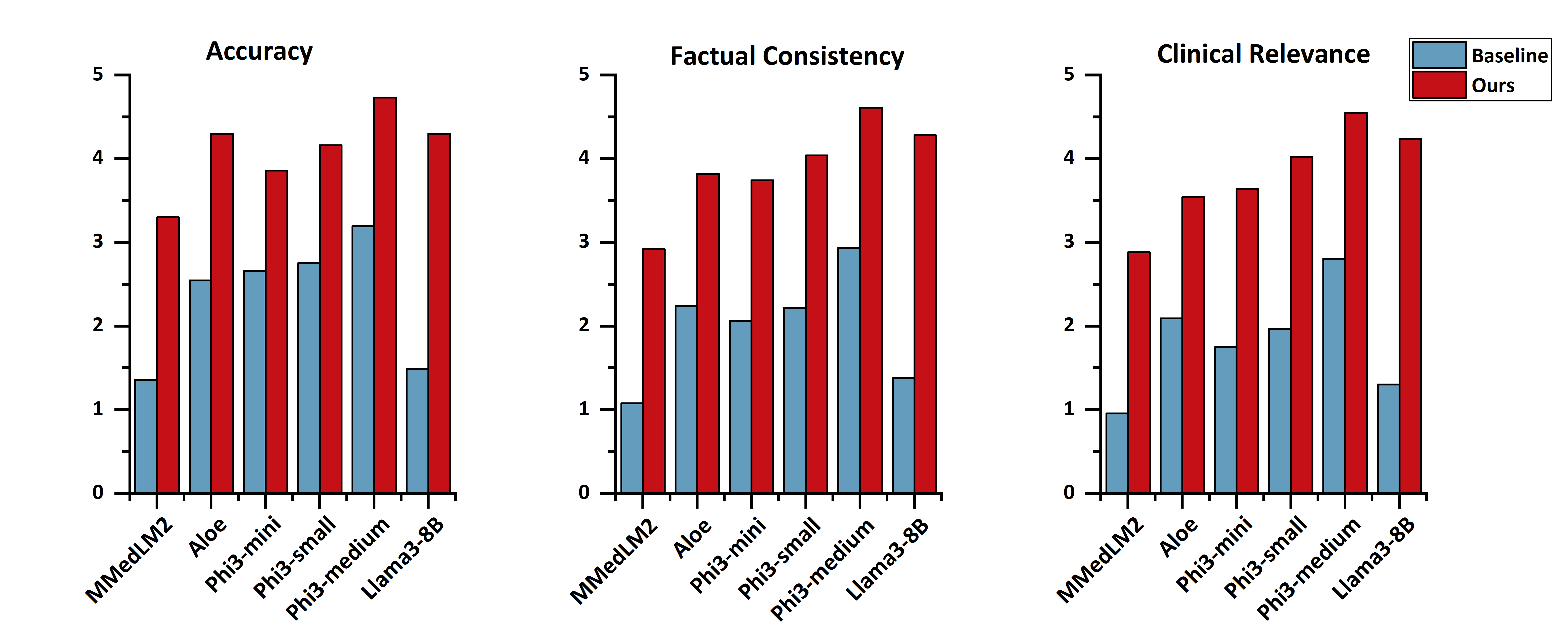}
  \caption{Human evaluation comparison of error correction on baseline versus our proposed staged proofreading inference with dual-knowledge infusion framework.}
  \label{fig:human_eval1}
\end{figure} 


To provide a more comprehensive validation of our method’s clinical relevance, we conducted a human evaluation with two practicing radiologists on 50 representative cases across six model architectures, comparing baseline inference with our staged proofreading framework using \ref{fig:human_eval1} five-point Likert scale (1–5). Figure 3 shows that our method consistently and significantly outperforms all baselines across three key dimensions—accuracy, factual consistency, and clinical relevance. The improvements are most pronounced for models such as Phi3-medium and Llama3-8B, where both factual correctness and clinical interpretability were notably enhanced, while MMedLM2 demonstrates particularly large gains in factual consistency and relevance to real-world diagnostic practice. These results highlight the essential role of expert evaluation in assessing medical text generation systems and confirm that our framework produces corrections that are not only linguistically precise but also clinically meaningful, trustworthy, and aligned with radiologists’ reasoning processes.

\subsection{Ablation Study}

\begin{table}[h]
\centering
\small
\begin{tabular}{lcccc}
\toprule
\textbf{Method} & \textbf{Det.} & \textbf{Loc.} & \textbf{Corr.} & \textbf{$\Delta$ Avg.} \\
& \textbf{(\%)} & \textbf{(\%)} & \textbf{(NLG)} & \textbf{(\%)} \\
\midrule
Baseline & 37.79 & 37.29 & 94.29 & - \\
+ MKGD & 37.92 & 36.99 & 94.49 & \textcolor{gray}{+0.01} \\
+ EXKR & 58.14 & 50.21 & 94.38 & \textcolor{ForestGreen}{+11.12} \\
+ MKGD + EXKR (Ours) & \textbf{62.27} & \textbf{53.14} & \textbf{94.43} & \textcolor{ForestGreen}{\textbf{+13.49}} \\
\bottomrule
\end{tabular}
\caption{Ablation Study on Llama3-8B for the three tasks. $\Delta$ Avg. represents average improvement across three tasks.}
\label{tab:llama3_ablation}
\end{table}

Table \ref{tab:llama3_ablation} presents a focused ablation study on Llama3-8B to understand the individual contributions of our framework components. The results reveal distinct patterns across different tasks and highlight the synergistic effects of our dual-knowledge infusion approach. MKGD alone provides minimal improvements across all tasks, with changes ranging from -0.30\% to +0.20\%, indicating that structured knowledge representation alone is insufficient. However, EXKR demonstrates substantial improvements in detection (+20.35\%) and localization (+12.92\%) but minimal impact on correction (+0.09\%), suggesting that external knowledge retrieval is particularly valuable for error identification but less critical for generating corrections. The combination of MKGD + EXKR yields the best performance across all tasks (+24.48\%, +15.85\%, +0.14\%), with the synergistic effect being most pronounced in detection and localization tasks. This pattern demonstrates that while EXKR drives the primary improvements, MKGD provides essential structural guidance that enables EXKR to work more effectively, particularly for tasks requiring precise understanding of medical entity relationships and clinical context.

\section{Conclusion}

In summary, this work addressed the challenge of automated error detection and correction in radiology reports through a framework combining staged proofreading inference with dual-knowledge infusion. Our approach decomposes error correction into three phases—detection, localization, and correction—while integrating structured medical knowledge through medical knowledge graph distillation (MKGD) with external knowledge retrieval (EXKR).

Evaluations across six language models demonstrate improvements: up to 31.56\% enhancement in error detection accuracy, with medical domain models showing notable gains. Our structured approach also achieves efficiency improvements, reducing processing time by up to 37.4\% while maintaining performance. Human evaluation by practicing radiologists indicates enhanced factual consistency and clinical relevance compared to existing approaches. Our ablation study shows that while MKGD provides structural guidance, EXKR contributes to the primary performance improvements, with their combination enabling error correction across tasks.

The framework transforms general-purpose language models into medical quality assurance tools without requiring expensive domain-specific fine-tuning. This addresses critical challenges in clinical practice: high error rates in radiological examinations, the growing gap between examination volumes and specialist availability, and the need for transparent, validatable AI systems that integrate with clinical workflows. Our approach provides automated proofreading that enhances clinical expertise, benefiting busy clinical environments and resource-constrained healthcare facilities where comprehensive manual review can be challenging. The principles established—staged inference modeling, dual-knowledge integration, and transparent reasoning—offer a foundation for developing AI systems that address healthcare documentation challenges while maintaining the interpretability and safety standards needed for clinical practice.

Despite these promising results, our work still has several limitations that point to future research directions: (1) extending the framework to multilingual and cross-specialty medical reports; (2) designing adaptive knowledge retrieval that evolves with clinical feedback; and (3) incorporating multimodal inputs—including images, previous reports, and patient history—to achieve more holistic error detection.

\section{Acknowledgments}
This research was supported by the Health and Medical Research Fund [Grant Nos. 19200911 and 21223141] and the National Natural Science Foundation of China Young Scientists Fund [Grant No. 82303957]. It also received support from the UK’s Engineering and Physical Sciences Research Council (EPSRC; UKRI2701: PAIR: Building a Cloneable Pipeline for Utilizing Foundation AI on EHRs), the Medical Research Council (MR/S004149/1, MR/X030075/1), and the British Council (Facilitating Better Urology Care With Effective and Fair Use of Artificial Intelligence—A Partnership Between UCL and Shanghai Jiao Tong University School of Medicine). HW’s role in this research was partially funded by the Legal \& General Group through a research grant to establish the independent Advanced Care Research Centre at the University of Edinburgh. The funders had no role in the conduct of the study, data interpretation, or the decision to submit this work for publication. We sincerely thank all funding agencies for their support.

\bibliography{aaai2026}

@inproceedings{abacha2023empirical,
  title={An empirical study of clinical note generation from doctor-patient encounters},
  author={Abacha, Asma Ben and Yim, Wen-wai and Fan, Yadan and Lin, Thomas},
  booktitle={Proceedings of the 17th Conference of the European Chapter of the Association for Computational Linguistics},
  pages={2291--2302},
  year={2023}
}

@article{wu2022survey,
  title={A survey on clinical natural language processing in the United Kingdom from 2007 to 2022},
  author={Wu, Honghan and Wang, Minhong and Wu, Jinge and Francis, Farah and Chang, Yun-Hsuan and Shavick, Alex and Dong, Hang and Poon, Michael TC and Fitzpatrick, Natalie and Levine, Adam P and others},
  journal={NPJ digital medicine},
  volume={5},
  number={1},
  pages={186},
  year={2022},
  publisher={Nature Publishing Group UK London}
}

@article{zhou2023survey,
  title={A survey of large language models in medicine: Progress, application, and challenge},
  author={Zhou, Hongjian and Gu, Boyang and Zou, Xinyu and Li, Yiru and Chen, Sam S and Zhou, Peilin and Liu, Junling and Hua, Yining and Mao, Chengfeng and Wu, Xian and others},
  journal={arXiv preprint arXiv:2311.05112},
  year={2023}
}

@article{jain2021radgraph,
  title={Radgraph: Extracting clinical entities and relations from radiology reports},
  author={Jain, Saahil and Agrawal, Ashwin and Saporta, Adriel and Truong, Steven QH and Duong, Du Nguyen and Bui, Tan and Chambon, Pierre and Zhang, Yuhao and Lungren, Matthew P and Ng, Andrew Y and others},
  journal={arXiv preprint arXiv:2106.14463},
  year={2021}
}

@inproceedings{lin2004rouge,
  title={Rouge: A package for automatic evaluation of summaries},
  author={Lin, Chin-Yew},
  booktitle={Text summarization branches out},
  pages={74--81},
  year={2004}
}

@article{zhang2019bertscore,
  title={Bertscore: Evaluating text generation with bert},
  author={Zhang, Tianyi and Kishore, Varsha and Wu, Felix and Weinberger, Kilian Q and Artzi, Yoav},
  journal={arXiv preprint arXiv:1904.09675},
  year={2019}
}

@article{sellam2020bleurt,
  title={BLEURT: Learning robust metrics for text generation},
  author={Sellam, Thibault and Das, Dipanjan and Parikh, Ankur P},
  journal={arXiv preprint arXiv:2004.04696},
  year={2020}
}

@inproceedings{abacha2023investigation,
  title={An investigation of evaluation methods in automatic medical note generation},
  author={Abacha, Asma Ben and Yim, Wen-wai and Michalopoulos, George and Lin, Thomas},
  booktitle={Findings of the Association for Computational Linguistics: ACL 2023},
  pages={2575--2588},
  year={2023}
}

@article{johnson2019mimic,
  title={MIMIC-CXR, a de-identified publicly available database of chest radiographs with free-text reports},
  author={Johnson, Alistair EW and Pollard, Tom J and Berkowitz, Seth J and Greenbaum, Nathaniel R and Lungren, Matthew P and Deng, Chih-ying and Mark, Roger G and Horng, Steven},
  journal={Scientific data},
  volume={6},
  number={1},
  pages={317},
  year={2019},
  publisher={Nature Publishing Group UK London}
}

@article{rajpurkar2022ai,
  title={AI in health and medicine},
  author={Rajpurkar, Pranav and Chen, Emma and Banerjee, Oishi and Topol, Eric J},
  journal={Nature medicine},
  volume={28},
  number={1},
  pages={31--38},
  year={2022},
  publisher={Nature Publishing Group US New York}
}

@article{qiu2024towards,
  title={Towards Building Multilingual Language Model for Medicine},
  author={Qiu, Pengcheng and Wu, Chaoyi and Zhang, Xiaoman and Lin, Weixiong and Wang, Haicheng and Zhang, Ya and Wang, Yanfeng and Xie, Weidi},
  journal={arXiv preprint arXiv:2402.13963},
  year={2024}
}

@article{gururajan2024aloe,
  title={Aloe: A Family of Fine-tuned Open Healthcare LLMs},
  author={Gururajan, Ashwin Kumar and Lopez-Cuena, Enrique and Bayarri-Planas, Jordi and Tormos, Adrian and Hinjos, Daniel and Bernabeu-Perez, Pablo and Arias-Duart, Anna and Martin-Torres, Pablo Agustin and Urcelay-Ganzabal, Lucia and Gonzalez-Mallo, Marta and others},
  journal={arXiv preprint arXiv:2405.01886},
  year={2024}
}

@article{abdin2024phi3,
  title={Phi-3 technical report: A highly capable language model locally on your phone},
  author={Abdin, Marah and Jacobs, Sam Ade and Awan, Ammar Ahmad and Aneja, Jyoti and Awadallah, Ahmed and Awadalla, Hany and Bach, Nguyen and Bahree, Amit and Bakhtiari, Arash and Behl, Harkirat and others},
  journal={arXiv preprint arXiv:2404.14219},
  year={2024}
}

@article{e5,
  title={Text embeddings by weakly-supervised contrastive pre-training},
  author={Wang, Liang and Yang, Nan and Huang, Xiaolong and Jiao, Binxing and Yang, Linjun and Jiang, Daxin and Majumder, Rangan and Wei, Furu},
  journal={arXiv preprint arXiv:2212.03533},
  year={2022}
}

@article{kasalak2023work,
  title={Work overload and diagnostic errors in radiology},
  author={Kasalak, {\"O}mer and Alnahwi, Hasan and Toxopeus, Rob and Pennings, Jan Pieter and Yakar, Derya and Kwee, Thomas C},
  journal={European Journal of Radiology},
  volume={167},
  pages={111032},
  year={2023},
  publisher={Elsevier}
}

@article{brady2017error,
  title={Error and discrepancy in radiology: inevitable or avoidable?},
  author={Brady, Adrian P},
  journal={Insights into Imaging},
  volume={8},
  number={1},
  pages={171--182},
  year={2017},
  publisher={Springer}
}

@article{fang2024llmcl,
  title={LLMCL-GEC: Advancing Grammatical Error Correction with LLM-Driven Curriculum Learning},
  author={Fang, Tao and Wong, Derek F and Zhang, Liang and Jin, Kaiyu and Zhang, Qiyu and Li, Tingting and Hou, Jingbo and Chao, Lidia S},
  journal={arXiv preprint arXiv:2412.12541},
  year={2024}
}

@article{kamoi2024evaluating,
  title={Evaluating LLMs at Detecting Errors in LLM Responses},
  author={Kamoi, Ryo and Das, Sarkar Snigdha Sarathi and Lou, Renze and Ahn, Jihyun Janice and Zhao, Yilun and Lu, Xiaoxin and Zhang, Nan and Zhang, Yue and Zhang, Rui Hao and Vummanthala, Sravya Reddy and others},
  journal={arXiv preprint arXiv:2404.03602},
  year={2024}
}

@inproceedings{ben2024overview,
  title={Overview of the MEDIQA-CORR 2024 Shared Task on Medical Error Detection and Correction},
  author={Ben Abacha, Asma and Yim, Wen-wai and Fu, Vivian and Sun, Zhaoyi and Xia, Fei and Yetisgen, Meliha},
  booktitle={Proceedings of the 6th Clinical Natural Language Processing Workshop},
  pages={1--10},
  year={2024},
  address={Mexico City, Mexico},
  publisher={Association for Computational Linguistics}
}

@article{lazaridou2021mind,
  title={Mind the gap: Assessing temporal generalization in neural language models},
  author={Lazaridou, Anna and Kuncoro, Adhiguna and Gribovskaya, Elena and Agrawal, Devang and Liska, Adam and Mairesse, Fran{\c{c}}ois and Blunsom, Phil},
  journal={Advances in Neural Information Processing Systems},
  volume={34},
  pages={29348--29363},
  year={2021}
}

@article{singhal2023large,
  title={Large language models encode clinical knowledge},
  author={Singhal, Karan and Azizi, Shekoofeh and Tu, Tao and Mahdavi, S Sara and Wei, Jason and Chung, Hyung Won and Scales, Nathan and Tanwani, Ajay and Cole-Lewis, Heather and Pfohl, Stephen and others},
  journal={Nature},
  volume={620},
  number={7972},
  pages={172--180},
  year={2023},
  publisher={Nature Publishing Group}
}

@article{collins2015new,
  title={A new initiative on precision medicine},
  author={Collins, Francis S and Varmus, Harold},
  journal={New England Journal of Medicine},
  volume={372},
  number={9},
  pages={793--795},
  year={2015},
  publisher={Massachusetts Medical Society}
}

@article{topol2019high,
  title={High-performance medicine: the convergence of human and artificial intelligence},
  author={Topol, Eric J},
  journal={Nature Medicine},
  volume={25},
  number={1},
  pages={44--56},
  year={2019},
  publisher={Nature Publishing Group}
}

@article{kruskal2011changes,
  title={Changes to the reporting radiologist workforce: implications for radiology residents and practicing radiologists},
  author={Kruskal, Jonathan B and Anderson, Stephan and Yam, Cho-Sun and Sosna, Jacob},
  journal={Radiology},
  volume={259},
  number={3},
  pages={616--621},
  year={2011},
  publisher={Radiological Society of North America}
}

@article{berlin2007malpractice,
  title={Malpractice issues in radiology: perceptual and cognitive errors},
  author={Berlin, Leonard},
  journal={Seminars in Ultrasound, CT and MRI},
  volume={28},
  number={3},
  pages={170--177},
  year={2007},
  publisher={Elsevier}
}

@article{sendak2020human,
  title={Human factors that influence the adoption of a sepsis clinical decision support system},
  author={Sendak, Mark P and Balu, Suresh and Schulman, Kevin A},
  journal={Applied Clinical Informatics},
  volume={11},
  number={3},
  pages={424--431},
  year={2020},
  publisher={Georg Thieme Verlag KG}
}

@article{cabitza2017unintended,
  title={Unintended consequences of machine learning in medicine},
  author={Cabitza, Federico and Rasoini, Raffaele and Gensini, Gian Franco},
  journal={JAMA},
  volume={318},
  number={6},
  pages={517--518},
  year={2017},
  publisher={American Medical Association}
}

@article{sunshine2006radiology,
  title={Radiology groups' workload in relative value units and factors affecting productivity},
  author={Sunshine, Jonathan H and Burkhardt, John H},
  journal={Radiology},
  volume={240},
  number={3},
  pages={504--515},
  year={2006},
  publisher={Radiological Society of North America}
}

@article{huang2019clinicalbert,
  title={ClinicalBERT: Modeling clinical notes and predicting hospital readmission},
  author={Huang, Kexin and Altosaar, Jaan and Ranganath, Rajesh},
  journal={arXiv preprint arXiv:1904.05342},
  year={2019}
}

@article{wang2018clinical,
  title={Clinical information extraction applications: a literature review},
  author={Wang, Yanshan and Wang, Liwei and Rastegar-Mojarad, Majid and Moon, Sungrim and Shen, Feichen and Afzal, Naveed and Liu, Sijia and Zeng, Yuqun and Mehrabi, Saeed and Sohn, Sunghwan and others},
  journal={Journal of Biomedical Informatics},
  volume={77},
  pages={34--49},
  year={2018},
  publisher={Elsevier}
}

@article{ng2020imaging,
  title={Imaging profile of the COVID-19 infection: radiologic findings and literature review},
  author={Ng, Ming-Yen and Lee, Elaine YP and Yang, Jin and Yang, Fangfang and Li, Xia and Wang, Hongxia and Lui, Macy Mei-sze and Lo, Christine Shing-Yen and Leung, Barry and Khong, Pek-Lan and others},
  journal={Radiology: Cardiothoracic Imaging},
  volume={2},
  number={1},
  pages={e200034},
  year={2020},
  publisher={Radiological Society of North America}
}

@article{pinto2018speech,
  title={Speech recognition in radiology: how to implement a successful solution},
  author={Pinto dos Santos, Daniel and Hempel, Julia-Marie and Mildenberger, Peter and Klöckner, Roman and Persigehl, Thorsten},
  journal={Insights into Imaging},
  volume={9},
  number={6},
  pages={891--895},
  year={2018},
  publisher={Springer}
}

@article{sun2025generative,
 title={Generative large language models trained for detecting errors in radiology reports},
 author={Sun, Cong and Teichman, Kurt and Zhou, Yiliang and Critelli, Brian and Nauheim, David and Keir, Graham and Wang, Xindi and others},
 journal={Radiology},
 volume={315},
 number={2},
 pages={e242575},
 year={2025},
 publisher={Radiological Society of North America}
}

@article{wu2024hybrid,
  title={A hybrid framework with large language models for rare disease phenotyping},
  author={Wu, Jinge and Dong, Hang and Li, Zexi and Wang, Haowei and Li, Runci and Patra, Arijit and Dai, Chengliang and Ali, Waqar and Scordis, Phil and Wu, Honghan},
  journal={BMC Medical Informatics and Decision Making},
  volume={24},
  number={1},
  pages={289},
  year={2024},
  publisher={Springer}
}

@article{wu2024radbartsum,
  title={RadBARTsum: Domain Specific Adaption of Denoising Sequence-to-Sequence Models for Abstractive Radiology Report Summarization},
  author={Wu, Jinge and Hasan, Abul and Wu, Honghan},
  journal={arXiv preprint arXiv:2406.03062},
  year={2024}
}

@article{wu2024hallucination,
  title={Hallucination benchmark in medical visual question answering},
  author={Wu, Jinge and Kim, Yunsoo and Wu, Honghan},
  journal={arXiv preprint arXiv:2401.05827},
  year={2024}
}

@article{wu2023exploring,
  title={Exploring multimodal large language models for radiology report error-checking},
  author={Wu, Jinge and Kim, Yunsoo and Keller, Eva C and Chow, Jamie and Levine, Adam P and Pontikos, Nikolas and Ibrahim, Zina and Taylor, Paul and Williams, Michelle C and Wu, Honghan},
  journal={arXiv preprint arXiv:2312.13103},
  year={2023}
}

\end{document}